\documentclass[conference]{IEEEtran}
\IEEEoverridecommandlockouts
\usepackage[utf8]{inputenc}
\usepackage[T1]{fontenc}
\usepackage{cite}
\usepackage{amsmath,amssymb,amsfonts}
\usepackage{algorithmic}
\usepackage{graphicx}
\usepackage{textcomp}
\usepackage{xcolor}
\usepackage{comment}
\usepackage{subcaption}
\usepackage[normalem]{ulem}
\usepackage{xurl}
\def\BibTeX{{\rm B\kern-.05em{\sc i\kern-.025em b}\kern-.08em
    T\kern-.1667em\lower.7ex\hbox{E}\kern-.125emX}}
\begin{document}

\title{SkelCap: Automated Generation of Descriptive Text from Skeleton Keypoint Sequences}

\author{
    \IEEEauthorblockN{Ali Emre Keskin}
    \IEEEauthorblockA{
        \textit{Computer Engineering Department} \\
        \textit{Hacettepe University} \\
        Ankara, Turkey \\
        aliemrekskn@gmail.com
    }
    \and
    \IEEEauthorblockN{Hacer Yalim Keles}
    \IEEEauthorblockA{
        \textit{Computer Engineering Department} \\
        \textit{Hacettepe University} \\
        Ankara, Turkey \\
        hacerkeles@cs.hacettepe.edu.tr
    }
}

\maketitle

\begin{abstract}

    Numerous sign language datasets exist, yet they typically cover only a limited selection of the thousands of signs used globally. Moreover, creating diverse sign language datasets is an expensive and challenging task due to the costs associated with gathering a varied group of signers. Motivated by these challenges, we aimed to develop a solution that addresses these limitations. In this context, we focused on textually describing body movements from skeleton keypoint sequences, leading to the creation of a new dataset. We structured this dataset around AUTSL, a comprehensive isolated Turkish sign language dataset. We also developed a baseline model, SkelCap, which can generate textual descriptions of body movements. This model processes the skeleton keypoints data as a vector, applies a fully connected layer for embedding, and utilizes a transformer neural network for sequence-to-sequence modeling. We conducted extensive evaluations of our model, including signer-agnostic and sign-agnostic assessments. The model achieved promising results, with a ROUGE-L score of 0.98 and a BLEU-4 score of 0.94 in the signer-agnostic evaluation. The dataset we have prepared, namely the AUTSL-SkelCap, will be made publicly available soon.

\end{abstract}


\section{\textbf{Introduction}}

According to the Ministry of Family and Social Services of Turkey, as of 2021, there were 213,553 individuals with hearing, language, or speech disabilities in Turkey \cite{eyhgm_istatistik_bulteni_temmuz2021}. For many of these individuals, sign language is the primary means of communication. Sign languages are not simple gestures or signed versions of spoken words; they are complete, independent linguistic systems. Therefore, despite advancements in electronic media, virtual reality, wearable electronics, and automatic simultaneous subtitling, sign languages remain indispensable for effective communication. Ongoing research aims to improve communication between sign language users and those who do not use sign language. This research is divided into three main categories: Sign Language Recognition (SLR), Sign Language Translation (SLT), and Sign Language Production (SLP).

Most research in sign language recognition, translation, and production relies on limited datasets. For instance, the BosphorusSign22k dataset \cite{ozdemir2020bosphorussign22k} includes only six unique signers, limiting its utility for signer-independent evaluations. Similarly, the RWTH-PHOENIX-Weather 2014T dataset \cite{koller2015continuous} is confined to weather forecast content, which represents only a narrow segment of typical human communication.

In recent years, deep neural networks have shown a strong ability to learn repeated patterns. Research in Sign Language Recognition (SLR) and Sign Language Translation (SLT) has incorporated these methods. However, there are challenges in applying these studies to real-world scenarios. A newer dataset, Elementary23 \cite{voskou2023new}, critiques the practical value of existing datasets and the SLT solutions evaluated with them.

To assess the real-world utility of existing datasets and solutions, consider AUTSL dataset \cite{sincan2020autsl}, one of the largest isolated Turkish sign language datasets in the literature. It contains 36,386 samples, which represent just 226 unique signs. Consequently, even the most advanced model trained on this dataset would only recognize up to 226 signs. While sign glosses do not directly correspond to spoken language words, for perspective, imagine relying on only 226 unique words in daily life. Compared to the 616,767 unique words in the Turkish language \cite{bozdemir2020ougrencilerin}, it becomes clear that current datasets are significantly limited for practical applications.

We initiated this study as part of a broader effort to address the issue of dataset inadequacy in sign language research. Our goal was not merely to enhance Sign Language Translation (SLT), but to develop a comprehensive solution by modeling the relationships between the movements involved in sign language and their textual equivalents. If a model can learn and generalize the motion characteristics of a sign sequence, it could potentially recognize or generate any sign (gloss), even those that are not included in the training corpus. This capability is particularly valuable given the challenges associated with creating data for each sign. Furthermore, if a model can accurately describe the content of a sign action, it suggests a fundamental understanding of the components that constitute a sign. This understanding can serve as a foundational model for further tuning and improving related tasks, such as sign recognition or translation.

This work primarily concentrates on two key areas. First, we focus on the development of a dataset that pairs textual descriptions with corresponding skeleton keypoint sequences, utilizing the AUTSL dataset. Second, we aim to develop a baseline model capable of generating textual descriptions from given skeletal sequences. This model will serve as a foundation for further research and improvement in the field of sign language processing.

The remainder of this paper is organized as follows: The next section covers Related Work, providing context and background for our study. This is followed by the Materials and Methods section, which details the creation of our dataset and the development of the proposed model. Subsequent sections present the Experimental Results and discuss their implications. The paper concludes with the Conclusion, summarizing our findings and suggesting directions for future research.

\section{\textbf{Related Work}}  \label{Related Works}

\subsection{\textbf{Action Recognition Datasets}}

Our study focuses on precisely recognizing and generating body movements, which existing datasets do not adequately address. Datasets such as the NTU RGB+D Dataset \cite{shahroudy2016ntu}, with 56 thousand video samples across 60 different action classes, and the Penn Action Dataset \cite{zhang2013actemes}, consisting of 2,326 consumer videos for 15 actions with keypoint annotations of moving body parts, are designed for broader action classifications. The MPII Human Pose Dataset \cite{andriluka20142d} features about 25,000 images from online videos annotated with 410 unique human activity labels, and the Charades Dataset \cite{sigurdsson2016hollywood} includes 9,848 videos for 157 action classes along with 27,847 textual descriptions. These descriptions often cover multiple aspects of a single video. The AVA Dataset \cite{gu2018ava} annotates 80 atomic visual actions in 430 movie clips of 15 minutes each, with a total of 1.62 million action labels annotated in both space and time.

In general, these datasets provide labels for broad actions, such as 'a chef cooking in the kitchen', which are too general for our needs. The detail we need involves more specific descriptions of body  highlighting the need for datasets that focus more on the specifics of individual body part and hand pose movements. This gap highlights the unique contribution of our work in developing a dataset and model that address this higher level of detail in action recognition.

\subsection{\textbf{Sign Language Recognition and Translation}}

Sign Language Recognition (SLR) tasks, as defined in existing literature, fall into two broad categories: isolated and continuous. Isolated SLR focuses on classifying each sign individually, without addressing the combination of signs into coherent sentences. Conversely, Continuous SLR involves the segmentation and classification of consecutive signs. Meanwhile, Sign Language Translation (SLT) contains the entire process of translating expressions from sign language to spoken language.

Our study does not directly address either SLR or SLT problem. However, we focus on these areas because the data modality used in SLR and SLT is similar to what we intend to use in our research. Understanding how these studies interpret and utilize their data will enhance our ability to effectively analyze similar data. In SLR, the input may be a video of sign sequences or a skeleton keypoint sequence, and the output is a single class for the recognized sign, employing sequence classification modeling. In contrast, SLT also takes a sign sequence video or skeleton keypoint sequence as input but outputs a sequence of spoken language tokens, using sequence-to-sequence modeling.

\textbf{\textit{Non-Transformer Era for SLR:}} In earlier studies like \cite{tur2019isolated} RNNs are used for sequence classification. They extract features with a Siamese Neural Network and concatenate them with depth information to use in the classification. To make the model more robust, in \cite{sincan2019isolated} Feature Pyramids and LSTM's are utilized. They get better test scores with smaller backbone feature extractor networks compared to the RNN based methods. While the methods keep getting better scores with bigger neural network models, the computation cost is also getting higher. The work in \cite{tur2021evaluation} points the efficiency of these systems. With a dimension reduction and HMM's they get comparable results with better computing efficiency.

\textbf{\textit{Skeleton Usage in SLR:}} Instead of using sign video directly, SPOTER\cite{bohavcek2022sign} makes the recognition based on the estimation of the pose of the human body in the form of 2D landmark positions. A robust pose normalization scheme is presented that processes hand poses in a separate local coordinate system regardless of body posture. These pose information are fed into a transformer encoder-decoder architecture for classification. A multi-modal approach by combining skeletal information with RGB and depth is proposed in SAM-SLR-v2\cite{jiang2021skeleton}. Feature extraction is performed in a total of 7 different modalities and the final classification is made with the global ensemble model.

\begin{figure*}[htbp]
    \centerline{\includegraphics[width=1\textwidth]{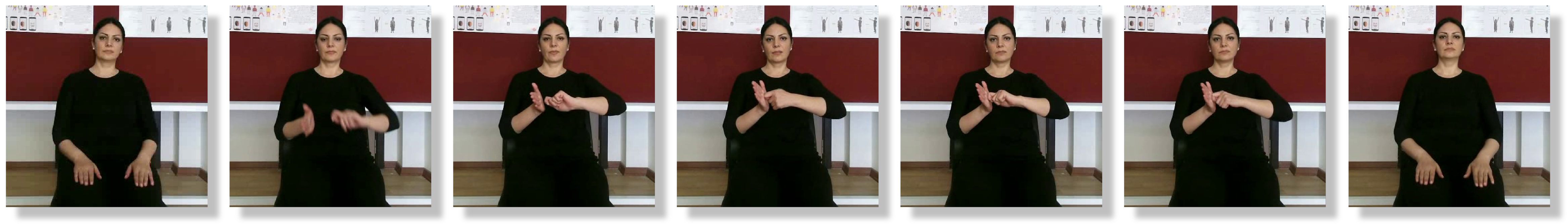}}
    \centerline{\includegraphics[width=1\textwidth]{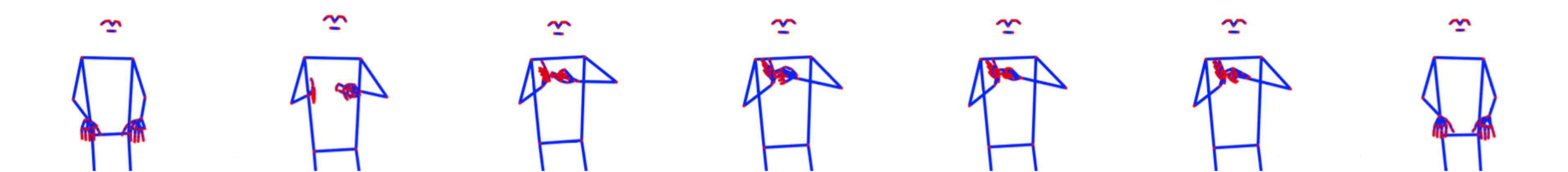}}
    \caption{A sample representative frame sequence from AUTSL dataset for "key" ( "anahtar" in Turkish ). Top: selected video frames, bottom: skeleton keypoint sequence.}
    \label{fig:anahtar}
\end{figure*}

\textbf{\textit{Textual Description Usage in SLR:}}
One of the works that inspired us is \cite{bilge2019zero}. In their study, they focus on the Zero-Shot Sign Language Recognition task. For this, they use textual descriptions of the signs, as in our study. As a methodology the body video sequence and the hand video sequence cut from the body video sequence are fed to 3D-CNNs separately. After that, separate Bidirectional LSTMs are used and finally the outputs are concatanated. On the other hand, textual descriptions are embedded with BERT. A final compatibility function is used to learn compatibility matrix between embeddings from video and text data. \cite{bilge2022towards} extends this work and includes discrete attributes like hand shape, hand location and movement type. \cite{ozcan2024enhancing}, also based on this work, enhances the framework with hand landmarks as well as hand video.

\textbf{\textit{Challenges of Sign Language Translation:}} As we move from SLR to SLT, we face the challenge of segmenting signs within a continuous stream. Furthermore, SLT requires the output to be a sequence of spoken language tokens, categorizing it as a sequence-to-sequence problem.

Successfully segmenting and identifying individual glosses does not equate to completing the translation process. As highlighted in the paper \cite{camgoz2018neural}, there is a distinct difference between Continuous Sign Language Recognition and Sign Language Translation. Sign languages are distinct from spoken languages; thus, recognizing a sequence of glosses does not automatically result in a spoken language translation. The sequence order can differ between glosses and spoken language words, and a single gloss may correspond to multiple words in spoken language.

\textbf{\textit{Sequence to Sequence Modeling in SLT:}} A pioneering study by \cite{camgoz2018neural} models the SLT problem using a sequence-to-sequence approach with RNNs. In their methodology, sign language videos are processed through a Convolutional Neural Network (CNN), which feeds into the encoder of their encoder-decoder architecture. The model is then trained to generate spoken language tokens from the decoder, effectively translating the sign language into spoken language.

In \cite{camgoz2020sign}, a method is proposed that does not require timing information for signs. This approach handles recognition and translation tasks jointly and end-to-end using a transformer-based encoder-decoder architecture. The encoder classifies sign language glosses, while the decoder generates spoken language texts.

\begin{figure}[htbp]
    \centerline{\includegraphics[width=0.35\textwidth]{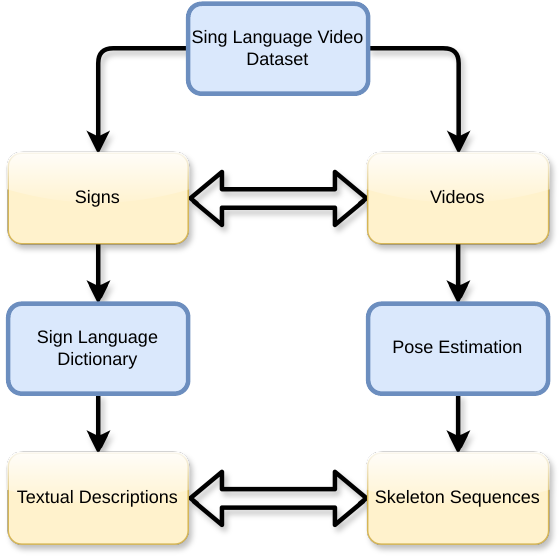}}
    \caption{Mapping between textual descriptions and skeleton sequences.}
    \label{fig:dataset-prep}
\end{figure}

\textbf{\textit{Skeleton Usage in SLT:}} Similar to the SLR problem, some approaches in SLT also utilize skeletons, as demonstrated in \cite{gan2021skeleton}, which introduces the Skeleton-Aware Neural Network (SANet) to leverage the distinctive and background invariant properties of skeletons. Additionally, a multi-channel transformer architecture that utilizes both video and skeleton information is proposed in \cite{camgoz2020multi}. Furthermore, the Hierarchical Deep Recurrent Fusion (HRF) framework, described in \cite{guo2019hierarchical}, is designed to effectively explore temporal cues in both video and skeleton sequences.

\section{\textbf{Materials and Methods}} \label{Proposed Method}

\subsection{\textbf{Dataset Preparation}}

We developed a framework to generate a dataset consisting of paired textual descriptions and skeleton keypoint sequences. This framework utilizes a sign language video dataset that includes isolated sign language videos with corresponding sign annotations. For each video-sign pair, we established two key mappings: one that links signs to textual descriptions and another that converts videos into skeleton sequences. These mappings allow us to directly associate textual descriptions with skeleton sequences, forming the core of our dataset. The relationships between these data types are illustrated in Figure \ref{fig:dataset-prep}.

\subsubsection{\textbf{Isolated Sign Language Video Dataset}}

Continuous sign language datasets, such as RWTH-PHOENIX-Weather 2014T \cite{koller2015continuous}, Content4All \cite{camgoz2021content4all}, and the SRF DSGS Daily News Broadcast \cite{jiang2023srf}, represent a significant advancement in sign language research by providing extensive video materials with sign-to-text translation annotations at the sentence level. These datasets offer a more realistic depiction of everyday sign language usage, facilitating the development of sequence-to-sequence models that capture the complex nuances of natural communication. However, for our purposes, these datasets introduce complexities due to the length and variability of each sentence, making them challenging for initial model training and evaluation.

Given the early stage of research in the proposed domain, starting with isolated sign datasets like AUTSL and MS-ASL\cite{vaezijoze2019ms-asl} is considered more feasible. These datasets, while focusing on isolated signs tagged with gloss labels for classification, allow for the exploration of how signs are performed as a sequence-to-sequence problem. This approach not only simplifies the initial modeling challenges but also serves as a foundational step towards understanding and processing more complex continuous sign language sequences. By initially focusing on isolated signs, we can develop robust models that can later be adapted and scaled to handle the complexities of continuous datasets.

Among various isolated sign language datasets, MS-ASL \cite{vaezijoze2019ms-asl}, a benchmark for American Sign Language, includes 1,000 different signs and features 200 signers, offering rich diversity in signs and signers. However, our focus on advancing Turkish sign language research led us to explore relevant local datasets. The BosphorusSign22k \cite{ozdemir2020bosphorussign22k} dataset, while containing a robust 744 different signs, only includes 6 signers. This limited signer variety makes it unsuitable for signer agnostic evaluations in our research.

Consequently, in this research we used AUTSL\cite{sincan2020autsl} dataset which is a large scale isolated Turkish sign language dataset containing total of 38,336 samples from 226 different signs performed by 43 different signers. Diversity in terms of signers in this dataset is a key part of signer agnostic description generation in our research. A minimal representation of a sample from AUTSL is shown in Figure \ref{fig:anahtar} Some of the key properties of these datasets are summarized in Table \ref{tab:propdataset}.

\begin{table}[htbp]
    \caption{Properties of isolated SLR datasets.}
    \begin{center}
        \begin{tabular}{|c|c|c|c|}
            \hline
            \textbf{dataset}      & \textbf{samples} & \textbf{signs} & \textbf{signers} \\ \hline
            AUTSL                 & 38,336           & 226            & 43               \\ \hline
            BosphorusSign22k      & 22,542           & 744            & 6                \\ \hline
            MS-ASL$^{\mathrm{a}}$ & 25,000           & 1000           & 200              \\ \hline
            \multicolumn{4}{l}{$^{\mathrm{a}}$Annotations in MS-ASL is in English, the others are in Turkish.}
        \end{tabular}
        \label{tab:propdataset}
    \end{center}
\end{table}

\subsubsection{\textbf{Mapping between Sings and Textual Descriptions}}

\begin{figure}[htbp]
    \centerline{
        \includegraphics[width=0.25\textwidth]{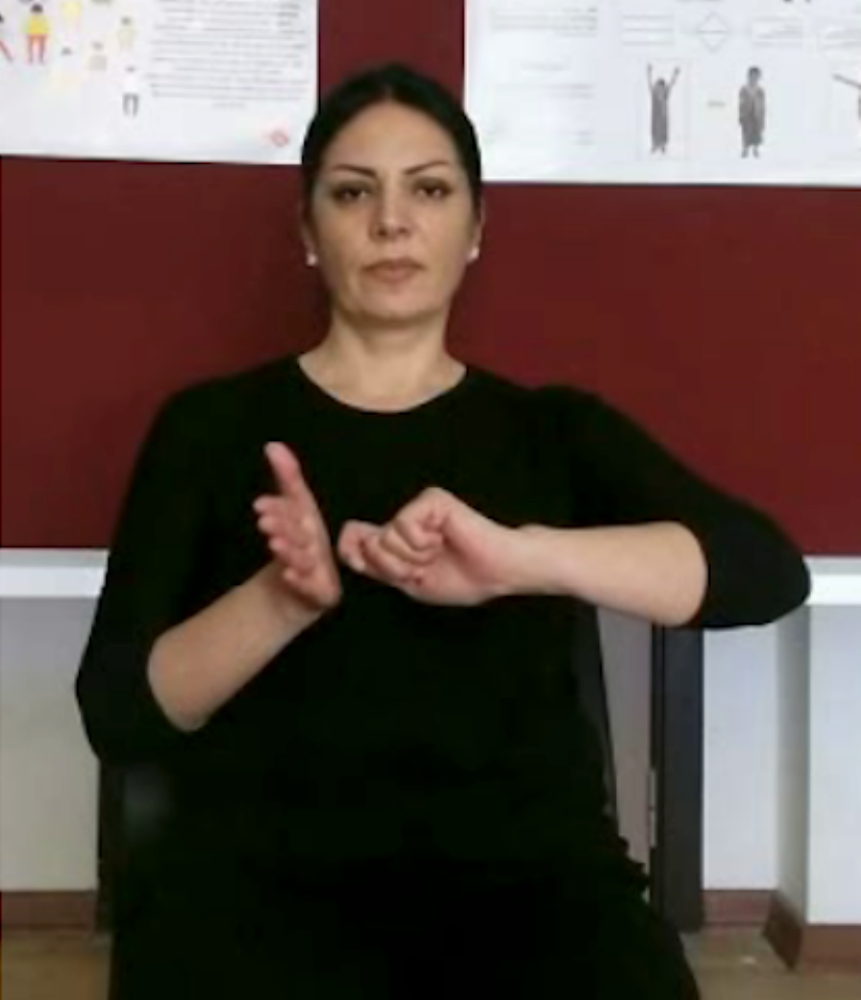}
    }
    \caption{Sample description of the 'key' sign from the Turkish Sign Language Dictionary (TSLD): 'The right hand is at chest level, shaped like a fist, with the index finger protruding forward and adjacent to the thumb (T hand). The right hand then rotates from the wrist to the right and left twice.' This image, adapted from a representative frame of the AUTSL dataset, illustrates how TSLD describes each sign.}
    \label{fig:tsl_sample}
\end{figure}

We used Turkish Sing Language Dictionary (TSLD) published by Republic of Turkey, Ministry of Education, General Directorate of Special Education and Guidance Services in 2015 \cite{güngör2015tsld}. The TSLD contains signs and their corresponding textual descriptions which describes how to perform the sign. A sample sign from the TSLD is "anahtar", which means "key" in English, is depicted in Figure \ref{fig:tsl_sample}. Sign "anahtar" is given with the description originally provided in Turkish as "Sağ el göğüs hizasında, yumruk biçiminde işaret parmağı öne doğru çıkıntılı ve başparmakla bitişiktir (T el). Sağ el bilekten iki kez sağa sola çevrilir.". The translated description to English is provided in the Figure \ref{fig:tsl_sample} caption.

The TSLD includes detailed information about categorical hand shapes and movements, which are essential for understanding sign execution in Turkish Sign Language (TSL). Table \ref{tab:handshapes} details specific hand shapes used in TSLD descriptions. In addition to hand shapes, hand movements within the dictionary are also depicted through symbolic representations; for example, a single arrow pointing right indicates a continuous straight movement of the hand. The dictionary defines a set of predefined movements as follows: 'Continuous straight movements', 'Continuation of straight movement in the same direction', 'Parallel movements in the same direction', 'Parallel movements in the opposite direction', 'All fingers start the movement open and finish it closed', 'All fingers start the movement closed and finish it open', 'Curved movements', and 'Circular movements'. These descriptions clarify how each movement is visually represented and interpreted within the context of the dictionary.

\begin{table*}[htbp]
    \centering
    \caption{Overview of TSL Hand Shapes}
    \label{tab:handshapes}
    \begin{tabular}{|l|l|}
        \hline
        \textbf{Hand Shape Name}                      & \textbf{Description}                                                                          \\ \hline
        C Hand                                        & Fingers and thumb form a semi-circle, resembling the letter C.                                \\ \hline
        T Hand                                        & Thumb extends horizontally across the palm, touching tips of the index and middle fingers.    \\ \hline
        L Hand                                        & Thumb and index finger extend to form a right angle.                                          \\ \hline
        U Hand                                        & Index and middle fingers are raised and held together, forming the letter U.                  \\ \hline
        P Hand                                        & Thumb is tucked between the extended index and middle fingers, with remaining fingers curled. \\ \hline
        V Hand                                        & Index and middle fingers are extended and separated, forming a V shape.                       \\ \hline
        Fingers open                                  & All fingers of the hand are spread apart and extended.                                        \\ \hline
        Fingers open and straight                     & All fingers are spread apart and fully extended in a straight position.                       \\ \hline
        Fingers open, straight, middle finger touches & Fingers open and straight; the middle finger extends to touch an object.                      \\ \hline
    \end{tabular}
    \caption*{\small Note: Descriptions are adapted from \cite{güngör2015tsld}, reflecting common conventions used in TSL interpretation.} 
\end{table*}

In the TSLD, certain signs are represented with multiple descriptions, reflecting variations in their execution. An example is the sign for "abla," which means "older sister" in English. TSLD provides two distinct descriptions for this sign: The first description details that "the index and middle fingers of the right hand are open and the other fingers are closed, with the palm facing the observer. Starting from the right edge of the chin, the open fingers are moved to rest on the right cheek." The second description explains that "the index and middle fingers are open and the other fingers closed, with the palm facing the observer, lifting the open fingers up after making contact with the chin." Such instances of multiple descriptions necessitate precise manual annotation in our research to ensure the selected sign representation aligns with the actual performance in the sign language video dataset.

\subsubsection{\textbf{Manual Annotation of the Hand Shapes and the Alternative Pronunciations}}

We manually stored the hand shape information of the signs. Firstly we assigned unique IDs to the hand shapes in the TSL Dictionary. Then we manually annotated every AUTSL sign's hand shape ID by looking the hand shape images in the TSL Dictionary. With this we have structured information of every AUTSL sign's hand shape information.

We inspected every AUTSL sign in the TSL Dictionary that has multiple pronunciations. We selected the correct alternative pronunciation by reading the TSL Dictionary descriptions and watching the AUTSL videos.

\begin{table}[htbp]
    \caption{Mapping of Missing Left Hand Landmarks to Nearest Body Landmarks}
    \begin{center}
        \begin{tabular}{| l | l | l | l |}
            \hline
            \textbf{\begin{tabular}[c]{@{}l@{}}left hand\\ landmark\\ index\end{tabular}} & \textbf{\begin{tabular}[c]{@{}l@{}}body\\ landmark\\ index\end{tabular}} & \textbf{\begin{tabular}[c]{@{}l@{}}left hand\\ landmark\\ name\end{tabular}}                                                                                                                                      & \textbf{\begin{tabular}[c]{@{}l@{}}body\\ landmark\\ name\end{tabular}} \\ \hline
            0                                                                             & 15                                                                       & wrist                                                                                                                                                                                                             & left\_wrist                                                             \\ \hline
            \begin{tabular}[c]{@{}l@{}}1\\ 2\\ 3\\ 4\end{tabular}                         & 21                                                                       & \begin{tabular}[c]{@{}l@{}}thumb\_cmc\\ thumb\_mcp\\ thumb\_ip\\ thumb\_tip\end{tabular}                                                                                                                          & left\_thumb                                                             \\ \hline
            \begin{tabular}[c]{@{}l@{}}5\\ 6\\ 7\\ 8\\ 9\\ 10\\ 11\\ 12\end{tabular}      & 19                                                                       & \begin{tabular}[c]{@{}l@{}}index\_finger\_mcp\\ index\_finger\_pip\\ index\_finger\_dip\\ index\_finger\_tip\\ middle\_finger\_mcp\\ middle\_finger\_pip\\ middle\_finger\_dip\\ middle\_finger\_tip\end{tabular} & left\_index                                                             \\ \hline
            \begin{tabular}[c]{@{}l@{}}13\\ 14\\ 15\\ 16\\ 17\\ 18\\ 19\\ 20\end{tabular} & 17                                                                       & \begin{tabular}[c]{@{}l@{}}ring\_finger\_mcp\\ ring\_finger\_pip\\ ring\_finger\_dip\\ ring\_finger\_tip\\ pinky\_finger\_mcp\\ pinky\_finger\_pip\\ pinky\_finger\_dip\\ pinky\_finger\_tip\end{tabular}         & left\_pinky                                                             \\ \hline
        \end{tabular}
        \label{tab:left_hand_body_mapping}
    \end{center}
\end{table}

\begin{table}[htbp]
    \caption{Mapping of Missing Right Hand Landmarks to Nearest Body Landmarks}
    \begin{center}
        \begin{tabular}{| l | l | l | l |}
            \hline
            \textbf{\begin{tabular}[c]{@{}l@{}}right hand\\ landmark\\ index\end{tabular}} & \textbf{\begin{tabular}[c]{@{}l@{}}body\\ landmark\\ index\end{tabular}} & \textbf{\begin{tabular}[c]{@{}l@{}}right hand\\ landmark\\ name\end{tabular}}                                                                                                                                     & \textbf{\begin{tabular}[c]{@{}l@{}}body\\ landmark\\ name\end{tabular}} \\
            \hline
            0                                                                              & 16                                                                       & wrist                                                                                                                                                                                                             & right\_wrist                                                            \\
            \hline
            \begin{tabular}[c]{@{}l@{}}1\\ 2\\ 3\\ 4\end{tabular}                          & 22                                                                       & \begin{tabular}[c]{@{}l@{}}thumb\_cmc\\ thumb\_mcp\\ thumb\_ip\\ thumb\_tip\end{tabular}                                                                                                                          & right\_thumb                                                            \\ \hline
            \begin{tabular}[c]{@{}l@{}}5\\ 6\\ 7\\ 8\\ 9\\ 10\\ 11\\ 12\end{tabular}       & 20                                                                       & \begin{tabular}[c]{@{}l@{}}index\_finger\_mcp\\ index\_finger\_pip\\ index\_finger\_dip\\ index\_finger\_tip\\ middle\_finger\_mcp\\ middle\_finger\_pip\\ middle\_finger\_dip\\ middle\_finger\_tip\end{tabular} & right\_index                                                            \\ \hline
            \begin{tabular}[c]{@{}l@{}}13\\ 14\\ 15\\ 16\\ 17\\ 18\\ 19\\ 20\end{tabular}  & 18                                                                       & \begin{tabular}[c]{@{}l@{}}ring\_finger\_mcp\\ ring\_finger\_pip\\ ring\_finger\_dip\\ ring\_finger\_tip\\ pinky\_finger\_mcp\\ pinky\_finger\_pip\\ pinky\_finger\_dip\\ pinky\_finger\_tip\end{tabular}         & right\_pinky                                                            \\ \hline
        \end{tabular}
        \label{tab:right_hand_body_mapping}
    \end{center}
\end{table}

\begin{figure}[htbp]
    \centerline{\includegraphics[width=0.4\textwidth]{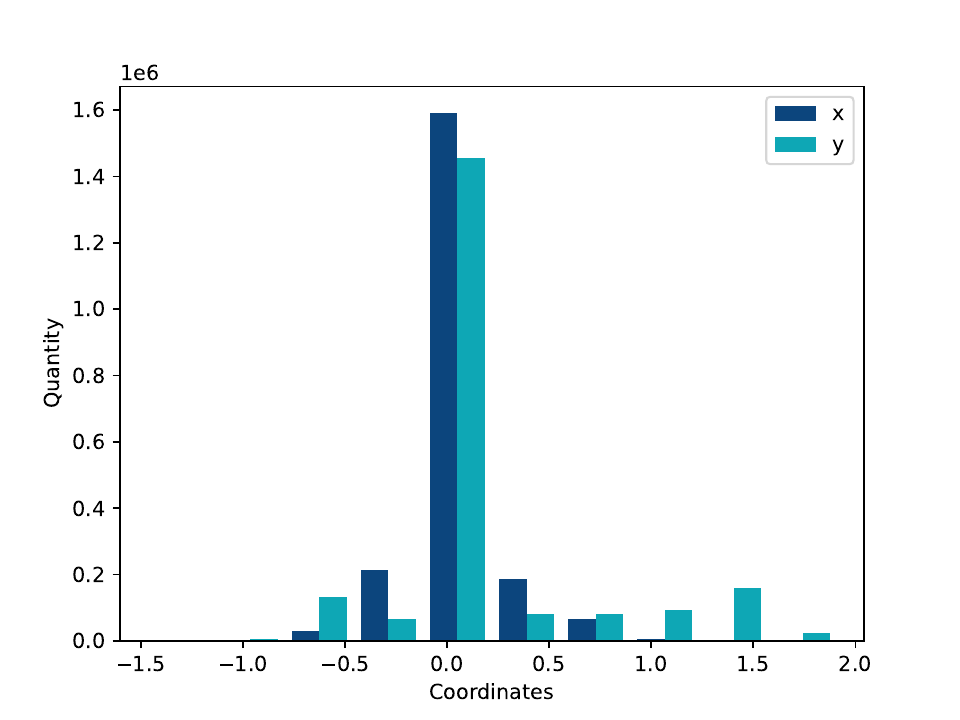}}
    \caption{Distribution of normalized skeleton points coordinates in x and y axes.}
    \label{fig:normalized_skeleton_coordinates_distribution}
\end{figure}

\subsubsection{\textbf{Mapping between Videos and Skeleton Sequences}}

\begin{figure*} [htbp]
    \centerline{\includegraphics[width=1\textwidth]{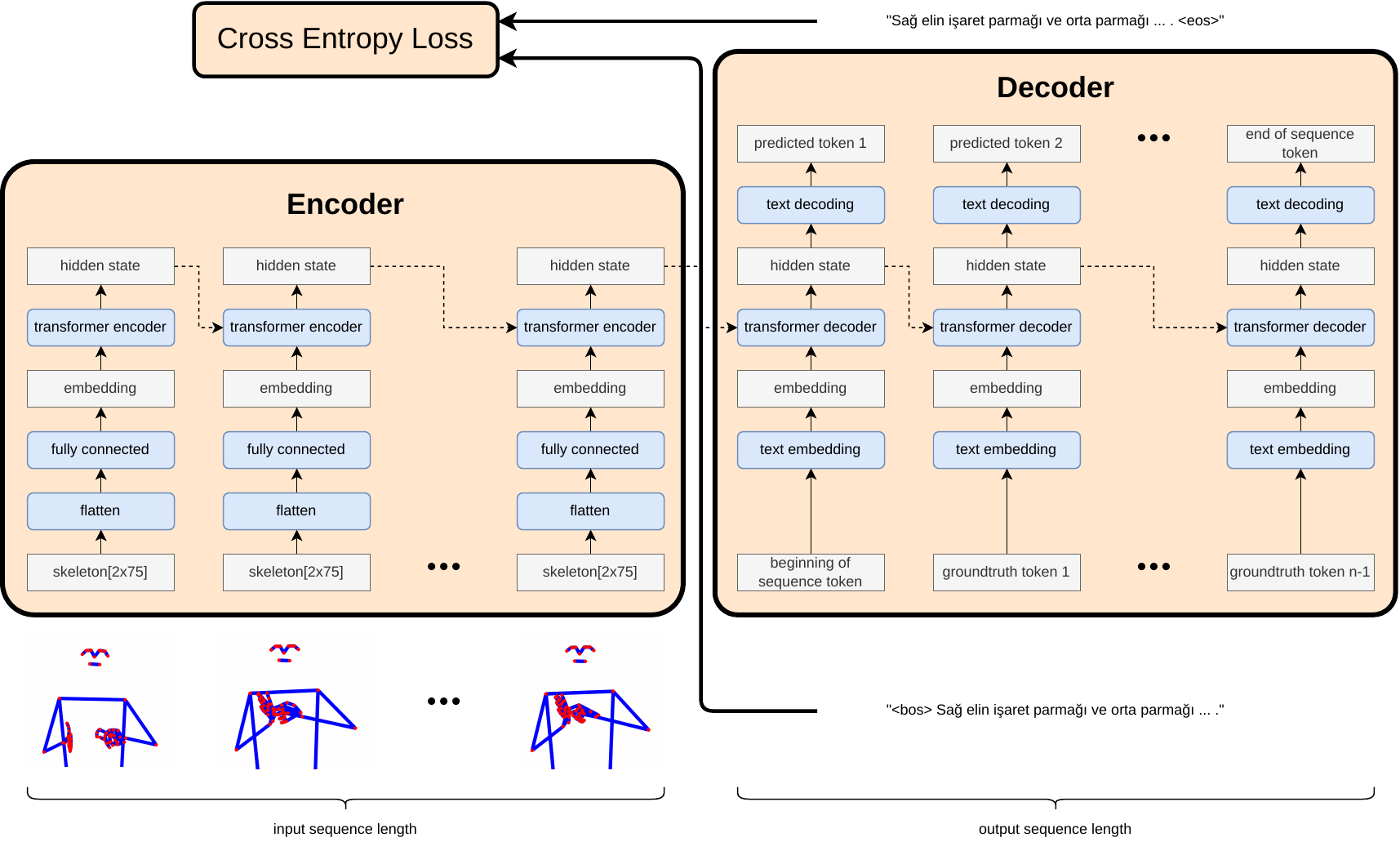}}
    \caption{Our sign skeleton sequence to text method.}
    \label{fig:skeleton_to_text_architecture}
\end{figure*}

Skeleton keypoints need to be detected from the videos in the dataset. At this stage, the OpenPose\cite{cao2017realtime} solution is examined. OpenPose can detect 135 key points, including human body, hand, face and foot key points, from a single image. However, there is a high memory requirement in the practical application phase. Another third-party solution, MediaPipe, is a solution that addresses many computer vision problems, not just the detection of skeleton keypoints. It uses the BlazePose GHUM\cite{grishchenko2022blazepose} method for the skeleton solution. We used MediaPipe to create the skeleton keypoint sequence part of the dataset. There are sub parts of the skeleton in the MediaPipe which are body, right hand and left hand. Graphical modelings of the skeleton can be examined in detail from \cite{mediapipe_pose_landmarker} for body pose landmarks and \cite{mediapipe_hand_landmarker} for hand landmarks.

Due to our reliance on a third-party solution for automatic skeleton extraction from videos, there are instances where the pose estimation may fail to retrieve complete skeleton information. These failures primarily affect the detection of hand and finger joint landmarks, while body landmarks are typically identified. In situations where finger joints cannot be detected, we assign to each undetected finger joint the value of the nearest hand landmark. The mappings from hand landmarks to body landmarks are detailed in Table \ref{tab:left_hand_body_mapping} for the left hand and Table \ref{tab:right_hand_body_mapping} for the right hand.

Sometimes, body landmarks may not be detected, leading to two scenarios: (1) the skeleton was successfully detected in images prior to the one where detection failed, and (2) the skeleton was not detected in any previous images. In the first scenario, we use the skeleton information from the preceding image for the image where the skeleton could not be detected. In the second scenario, where no prior detections are available, we assign a zero value to the coordinate information of all skeleton joints.

\subsubsection{\textbf{Skeleton Normalization}}

We applied spatial normalization to the detected skeletons following a two-step process. Initially, each skeleton was scaled so that the distance between the left and right shoulder points, indexed as 11 and 12 in \cite{mediapipe_pose_landmarker}, equals one unit. Subsequently, a translation adjustment was made to position the midpoint between these shoulder points at the origin. This normalization technique closely resembles the method described in \cite{stoll2018sign}.

The data distribution of the 2D points of the skeletons after normalization is shown in Fig. \ref{fig:normalized_skeleton_coordinates_distribution}.

\subsubsection{\textbf{Structured Storing of the Skeleton Sequences}}

After extracting the skeleton information and normalizing it, we have the skeleton points, scale and origin information. Skeleton points are a 150-element float array consisting of coordinates on the x and y axis. Scale is a one-dimensional scalar. Origin is a two-dimensional scalar. If desired, skeleton points and origin can be stored in three dimensions. We did not use third dimension data in this study.

\begin{table*}[htbp]
    \caption{Average metrics between dataset samples}
    \begin{center}
        \begin{tabular}{c||ccc|ccccc}
            \textbf{Setup}                  & \textbf{ROUGE-1} & \textbf{ROUGE-2} & \textbf{ROUGE-L} & \textbf{BLEU} & \textbf{BLEU-1} & \textbf{BLEU-2} & \textbf{BLEU-3} & \textbf{BLEU-4} \\ \hline \hline
            average between dataset samples & 0.32             & 0.14             & 0.26             & 0.08          & 0.33            & 0.11            & 0.05            & 0.02
        \end{tabular}
        \label{tab:dataset_metrics}
    \end{center}
\end{table*}

\begin{table*}[htbp]
    \caption{ROUGE values for text generation}
    \begin{center}
        \begin{tabular}{c||ccc|ccc}
                            &                  & \textbf{Train}   &                  &                  & \textbf{Test}    &                  \\
            \textbf{Setup}  & \textbf{ROUGE-1} & \textbf{ROUGE-2} & \textbf{ROUGE-L} & \textbf{ROUGE-1} & \textbf{ROUGE-2} & \textbf{ROUGE-L} \\ \hline \hline
            sign agnostic   & 0.81             & 0.66             & 0.8              & 0.52             & 0.27             & 0.49             \\ \hline
            signer agnostic & 0.99             & 0.98             & 0.99             & 0.98             & 0.96             & 0.98
        \end{tabular}
        \label{tab:rouge}
    \end{center}
\end{table*}

\begin{table*}[htbp]
    \caption{BLEU values for text generation}
    \begin{center}
        \begin{tabular}{c||ccccc|ccccc}
                            &               &                 & \textbf{Train}  &                 &                 &               &                 & \textbf{Test}   &                 &                 \\
            \textbf{Setup}  & \textbf{BLEU} & \textbf{BLEU-1} & \textbf{BLEU-2} & \textbf{BLEU-3} & \textbf{BLEU-4} & \textbf{BLEU} & \textbf{BLEU-1} & \textbf{BLEU-2} & \textbf{BLEU-3} & \textbf{BLEU-4} \\ \hline \hline
            sign agnostic   & 0.55          & 0.79            & 0.62            & 0.5             & 0.42            & 0.15          & 0.48            & 0.21            & 0.11            & 0.06            \\ \hline
            signer agnostic & 0.97          & 0.98            & 0.97            & 0.96            & 0.96            & 0.96          & 0.98            & 0.96            & 0.95            & 0.94
        \end{tabular}
        \label{tab:bleu}
    \end{center}
\end{table*}

\subsection{\textbf{SkelCap: Proposed Architecture}}

In our study, we approached to the sign skeleton sequence to caption generation problem as a sequence-to-sequence translation problem. One of the frequently used sequence-to-sequence deep models in NLP domain is the T5 model \cite{raffel2020exploring}. T5 is an encoder-decoder deep model that is pre-trained on a multiple mix of supervised and unsupervised tasks, with each task converted to a text-to-text format. This study is formed the backbone of the sequence-to-sequence modeling part of our study. We also used the T5's pre-trained weights for the text decoder.

Our proposed model architecture, depicted in Figure \ref{fig:skeleton_to_text_architecture}, processes the input sequence where each element consists of 75 skeleton keypoints, represented as a 75 x 2 tensor for their 2D positions. In our architecture, these keypoints are first transformed into vectors through a fully connected neural network (linear layer) to match the required input dimensions of the transformer. It's important to note that the input and output dimensions of the transformer network are identical, ensuring consistency in data handling throughout the model. During the encoding phase, these vectors, which represent the entire sequence, are fed into the transformer network. For decoding, the encoder's final hidden state, along with a start token, are used as inputs for the decoder block to generate the output tokens.

\subsubsection{\textbf{Training}}

We started with a pre-trained mT5-base weights provided in \cite{google/mt5-base}. Pre-trained model has the transformer model dimension of 768. Our skeleton embedding module, which is a linear layer, is initialized with random weights for input size of 150 and output size of 768. Since the pre-trained model's tokenizer has a vocabulary of size 250,112 for tokens the token embedding layer has input size of 250,112 and output size of 768. Size of the intermediate feed forward layer is 2,048. The number of transformer encoder layers and the number of transformer decoder layer is 12. The number of attention heads is also 12. Probability of the dropout layers to be zeroed is 0.1.

The output token is compared to the first word of the description text read from the dataset. For this comparison, a dictionary capacity sized classifier approach is applied and the loss function output is obtained. According to the output of this loss function, the whole network is passed backwards and the weights of the network are updated. This allows the model to learn to generate the given spoken language text in the dataset for the given sequence of skeleton keypoints.

We used Adam\cite{diederik2014adam} optimizer with a learning rate of 5e-5. The classical cross entropy loss is used for the token classification.

\section{\textbf{Experimental Results}} \label{Experimental Results}

We assessed our SkelCap model using ROUGE \cite{lin2004rouge} and BLEU \cite{papineni2002bleu} metrics. Initially, we calculated these metrics across all samples in the dataset to establish baseline distances for different samples, as shown in Table \ref{tab:dataset_metrics}. Subsequently, we implemented two distinct evaluation setups for training and testing. In the first setup, known as signer agnostic caption generation, we ensured that signers included in the training split did not appear in the test split. In the second setup, referred to as sign agnostic caption generation, we arranged the splits so that the same signs were not present in both training and test sets. The tarining and test scores for both setups are detailed in Table \ref{tab:rouge} for ROUGE scores and Table \ref{tab:bleu} for BLEU scores.

When we compare our model's performance with 0.26 ROUGE-L and 0.02 BLEU-4, which are the average metrics among the data in the dataset (Table \ref{tab:dataset_metrics}), we see that our base model is quite promising with the 0.98 ROUGE-L and 0.94 BLEU-4 values obtained for the signer agnostic setup on test data. The proposed baseline architecture generalizes well to the variations among signers.

On the other hand, as the results in the sign agnostic setup shows, our base model's capability to generate the caption for unseen signs is poor. During training, the model performs moderately well in this setting, with a BLEU overall score of 0.55. It does best with single words or short phrases, as shown by the highest score of 0.79 for BLEU-1. However, as the phrases get longer, the model struggles more, evidenced by the lower scores for longer n-grams like BLEU-4, which is only 0.42. When tested on unseen data (i.e. on a skeleton sequence obtained from an unseen sign), the performance drops significantly; the overall BLEU score is just 0.15, and it's even lower for longer sequences, at 0.06 for BLEU-4. This big drop in test scores signals the overfitting behaviour of the model; which means that while the model can handle various signs during training, it doesn't translate that learning effectively to new signs. This indicates a need for better methods that help the model generalize better from training to real-world applications. This task, related to zero-shot learning of sign descriptions, remains an open problem and is suggested as future work for interested researchers.

\section{\textbf{Conclusion}} \label{Conclusion}

We produced a dataset consisting of textual descriptions in spoken language and the skeleton keypoint sequences which these descriptions are performed. The dataset will be publicly available soon.

Our proposed skeleton to text model can produce promising textual descriptions from sign language skeleton sequences in a signer agnostic manner. Sign agnosticity for the sign language skeleton sequence to spoken language description text task is still open for further research.

\bibliographystyle{IEEEtran}
\bibliography{refs}

\begin{thebibliography}{10}
\providecommand{\url}[1]{#1}
\csname url@samestyle\endcsname
\providecommand{\newblock}{\relax}
\providecommand{\bibinfo}[2]{#2}
\providecommand{\BIBentrySTDinterwordspacing}{\spaceskip=0pt\relax}
\providecommand{\BIBentryALTinterwordstretchfactor}{4}
\providecommand{\BIBentryALTinterwordspacing}{\spaceskip=\fontdimen2\font plus
\BIBentryALTinterwordstretchfactor\fontdimen3\font minus \fontdimen4\font\relax}
\providecommand{\BIBforeignlanguage}[2]{{%
\expandafter\ifx\csname l@#1\endcsname\relax
\typeout{** WARNING: IEEEtran.bst: No hyphenation pattern has been}%
\typeout{** loaded for the language `#1'. Using the pattern for}%
\typeout{** the default language instead.}%
\else
\language=\csname l@#1\endcsname
\fi
#2}}
\providecommand{\BIBdecl}{\relax}
\BIBdecl

\bibitem{eyhgm_istatistik_bulteni_temmuz2021}
\BIBentryALTinterwordspacing
\emph{Engelli ve yasli istatistik Bulteni Temmuz 2021}.\hskip 1em plus 0.5em minus 0.4em\relax Republic of Turkey Ministry of Family and Social Services, 2021. [Online]. Available: \url{https://www.aile.gov.tr/media/88684/eyhgm_istatistik_bulteni_temmuz2021.pdf}
\BIBentrySTDinterwordspacing

\bibitem{ozdemir2020bosphorussign22k}
O.~{\"O}zdemir, A.~A. K{\i}nd{\i}ro{\u{g}}lu, N.~Cihan~Camgoz, and L.~Akarun, ``{BosphorusSign22k Sign Language Recognition Dataset},'' in \emph{Proceedings of the LREC2020 9th Workshop on the Representation and Processing of Sign Languages: Sign Language Resources in the Service of the Language Community, Technological Challenges and Application Perspectives}, 2020.

\bibitem{koller2015continuous}
O.~Koller, J.~Forster, and H.~Ney, ``Continuous sign language recognition: Towards large vocabulary statistical recognition systems handling multiple signers,'' \emph{Computer Vision and Image Understanding}, vol. 141, pp. 108--125, 2015.

\bibitem{voskou2023new}
A.~Voskou, K.~P. Panousis, H.~Partaourides, K.~Tolias, and S.~Chatzis, ``A new dataset for end-to-end sign language translation: The greek elementary school dataset,'' in \emph{Proceedings of the IEEE/CVF International Conference on Computer Vision}, 2023, pp. 1966--1975.

\bibitem{sincan2020autsl}
O.~M. Sincan and H.~Y. Keles, ``Autsl: A large scale multi-modal turkish sign language dataset and baseline methods,'' \emph{IEEE access}, vol.~8, pp. 181\,340--181\,355, 2020.

\bibitem{bozdemir2020ougrencilerin}
O.~BOZDEM{\.I}R, ``{\"O}{\u{g}}renc{\.i}ler{\.i}n g{\"o}r{\"u}{\c{s}}ler{\.i}ne g{\"o}re t{\"u}rk{\c{c}}e kel{\.i}me da{\u{g}}arci{\u{g}}inin azalmasi sorunu ve {\c{c}}{\"o}z{\"u}m {\"o}ner{\.i}ler{\.i},'' \emph{EKEV Akademi Dergisi}, no.~82, pp. 307--320, 2020.

\bibitem{shahroudy2016ntu}
A.~Shahroudy, J.~Liu, T.-T. Ng, and G.~Wang, ``Ntu rgb+ d: A large scale dataset for 3d human activity analysis,'' in \emph{Proceedings of the IEEE conference on computer vision and pattern recognition}, 2016, pp. 1010--1019.

\bibitem{zhang2013actemes}
W.~Zhang, M.~Zhu, and K.~G. Derpanis, ``From actemes to action: A strongly-supervised representation for detailed action understanding,'' in \emph{Proceedings of the IEEE international conference on computer vision}, 2013, pp. 2248--2255.

\bibitem{andriluka20142d}
M.~Andriluka, L.~Pishchulin, P.~Gehler, and B.~Schiele, ``2d human pose estimation: New benchmark and state of the art analysis,'' in \emph{Proceedings of the IEEE Conference on computer Vision and Pattern Recognition}, 2014, pp. 3686--3693.

\bibitem{sigurdsson2016hollywood}
G.~A. Sigurdsson, G.~Varol, X.~Wang, A.~Farhadi, I.~Laptev, and A.~Gupta, ``Hollywood in homes: Crowdsourcing data collection for activity understanding,'' in \emph{Computer Vision--ECCV 2016: 14th European Conference, Amsterdam, The Netherlands, October 11--14, 2016, Proceedings, Part I 14}.\hskip 1em plus 0.5em minus 0.4em\relax Springer, 2016, pp. 510--526.

\bibitem{gu2018ava}
C.~Gu, C.~Sun, D.~A. Ross, C.~Vondrick, C.~Pantofaru, Y.~Li, S.~Vijayanarasimhan, G.~Toderici, S.~Ricco, R.~Sukthankar \emph{et~al.}, ``Ava: A video dataset of spatio-temporally localized atomic visual actions,'' in \emph{Proceedings of the IEEE conference on computer vision and pattern recognition}, 2018, pp. 6047--6056.

\bibitem{tur2019isolated}
A.~O. Tur and H.~Y. Keles, ``Isolated sign recognition with a siamese neural network of rgb and depth streams,'' in \emph{IEEE EUROCON 2019-18th International Conference on Smart Technologies}.\hskip 1em plus 0.5em minus 0.4em\relax IEEE, 2019, pp. 1--6.

\bibitem{sincan2019isolated}
O.~M. Sincan, A.~O. Tur, and H.~Y. Keles, ``Isolated sign language recognition with multi-scale features using lstm,'' in \emph{2019 27th signal processing and communications applications conference (SIU)}.\hskip 1em plus 0.5em minus 0.4em\relax IEEE, 2019, pp. 1--4.

\bibitem{tur2021evaluation}
A.~O. Tur and H.~Y. Keles, ``Evaluation of hidden markov models using deep cnn features in isolated sign recognition,'' \emph{Multimedia Tools and Applications}, vol.~80, pp. 19\,137--19\,155, 2021.

\bibitem{bohavcek2022sign}
M.~Boh{\'a}{\v{c}}ek and M.~Hr{\'u}z, ``Sign pose-based transformer for word-level sign language recognition,'' in \emph{Proceedings of the IEEE/CVF winter conference on applications of computer vision}, 2022, pp. 182--191.

\bibitem{jiang2021skeleton}
S.~Jiang, B.~Sun, L.~Wang, Y.~Bai, K.~Li, and Y.~Fu, ``Skeleton aware multi-modal sign language recognition,'' in \emph{Proceedings of the IEEE/CVF conference on computer vision and pattern recognition}, 2021, pp. 3413--3423.

\bibitem{bilge2019zero}
Y.~C. Bilge, N.~Ikizler-Cinbis, and R.~G. Cinbis, ``Zero-shot sign language recognition: Can textual data uncover sign languages?'' \emph{arXiv preprint arXiv:1907.10292}, 2019.

\bibitem{bilge2022towards}
Y.~C. Bilge, R.~G. Cinbis, and N.~Ikizler-Cinbis, ``Towards zero-shot sign language recognition,'' \emph{IEEE transactions on pattern analysis and machine intelligence}, vol.~45, no.~1, pp. 1217--1232, 2022.

\bibitem{ozcan2024enhancing}
G.~S. {\"O}zcan, E.~S{\"U}MER, and Y.~C. B{\.I}LGE, ``Enhancing zero-shot learning based sign language recognition through hand landmarks and data augmentation,'' \emph{Journal of Millimeterwave Communication, Optimization and Modelling}, vol.~4, no.~1, pp. 17--20, 2024.

\bibitem{camgoz2018neural}
N.~C. Camgoz, S.~Hadfield, O.~Koller, H.~Ney, and R.~Bowden, ``Neural sign language translation,'' in \emph{Proceedings of the IEEE conference on computer vision and pattern recognition}, 2018, pp. 7784--7793.

\bibitem{camgoz2020sign}
N.~C. Camgoz, O.~Koller, S.~Hadfield, and R.~Bowden, ``Sign language transformers: Joint end-to-end sign language recognition and translation,'' in \emph{Proceedings of the IEEE/CVF conference on computer vision and pattern recognition}, 2020, pp. 10\,023--10\,033.

\bibitem{gan2021skeleton}
S.~Gan, Y.~Yin, Z.~Jiang, L.~Xie, and S.~Lu, ``Skeleton-aware neural sign language translation,'' in \emph{Proceedings of the 29th ACM International Conference on Multimedia}, 2021, pp. 4353--4361.

\bibitem{camgoz2020multi}
N.~C. Camgoz, O.~Koller, S.~Hadfield, and R.~Bowden, ``Multi-channel transformers for multi-articulatory sign language translation,'' in \emph{Computer Vision--ECCV 2020 Workshops: Glasgow, UK, August 23--28, 2020, Proceedings, Part IV 16}.\hskip 1em plus 0.5em minus 0.4em\relax Springer, 2020, pp. 301--319.

\bibitem{guo2019hierarchical}
D.~Guo, W.~Zhou, A.~Li, H.~Li, and M.~Wang, ``Hierarchical recurrent deep fusion using adaptive clip summarization for sign language translation,'' \emph{IEEE Transactions on Image Processing}, vol.~29, pp. 1575--1590, 2019.

\bibitem{camgoz2021content4all}
N.~C. Camg{\"o}z, B.~Saunders, G.~Rochette, M.~Giovanelli, G.~Inches, R.~Nachtrab-Ribback, and R.~Bowden, ``Content4all open research sign language translation datasets,'' in \emph{2021 16th IEEE International Conference on Automatic Face and Gesture Recognition (FG 2021)}.\hskip 1em plus 0.5em minus 0.4em\relax IEEE, 2021, pp. 1--5.

\bibitem{jiang2023srf}
Z.~Jiang, M.~M{\"u}ller, S.~Ebling, A.~Moryossef, and R.~Ribback, ``Srf dsgs daily news broadcast: video and original subtitle data,'' 2023.

\bibitem{vaezijoze2019ms-asl}
\BIBentryALTinterwordspacing
H.~Vaezi~Joze and O.~Koller, ``Ms-asl: A large-scale data set and benchmark for understanding american sign language,'' in \emph{The British Machine Vision Conference (BMVC)}, September 2019. [Online]. Available: \url{https://www.microsoft.com/en-us/research/publication/ms-asl-a-large-scale-data-set-and-benchmark-for-understanding-american-sign-language/}
\BIBentrySTDinterwordspacing

\bibitem{güngör2015tsld}
\BIBentryALTinterwordspacing
C.~Gungor, T.~Bagriacik, I.~Demirdogen, A.~Gunaydin, and V.~Karahan, \emph{Turkish Sign Language Dictionary}.\hskip 1em plus 0.5em minus 0.4em\relax Republic of Turkey, Ministry of Education, General Directorate of Special Education and Guidance Services, 2015. [Online]. Available: \url{https://orgm.meb.gov.tr/dosyalar/00012/tid_sozluk.pdf}
\BIBentrySTDinterwordspacing

\bibitem{cao2017realtime}
Z.~Cao, T.~Simon, S.-E. Wei, and Y.~Sheikh, ``Realtime multi-person 2d pose estimation using part affinity fields,'' in \emph{Proceedings of the IEEE conference on computer vision and pattern recognition}, 2017, pp. 7291--7299.

\bibitem{grishchenko2022blazepose}
I.~Grishchenko, V.~Bazarevsky, A.~Zanfir, E.~G. Bazavan, M.~Zanfir, R.~Yee, K.~Raveendran, M.~Zhdanovich, M.~Grundmann, and C.~Sminchisescu, ``Blazepose ghum holistic: Real-time 3d human landmarks and pose estimation,'' \emph{arXiv preprint arXiv:2206.11678}, 2022.

\bibitem{mediapipe_pose_landmarker}
\BIBentryALTinterwordspacing
``Pose landmark detection guide,'' 2023. [Online]. Available: \url{https://developers.google.com/mediapipe/solutions/vision/pose_landmarker}
\BIBentrySTDinterwordspacing

\bibitem{mediapipe_hand_landmarker}
\BIBentryALTinterwordspacing
``Hand landmarks detection guide,'' 2023. [Online]. Available: \url{https://developers.google.com/mediapipe/solutions/vision/hand_landmarker}
\BIBentrySTDinterwordspacing

\bibitem{stoll2018sign}
S.~Stoll, N.~C. Camg{\"o}z, S.~Hadfield, and R.~Bowden, ``Sign language production using neural machine translation and generative adversarial networks,'' in \emph{Proceedings of the 29th British Machine Vision Conference (BMVC 2018)}.\hskip 1em plus 0.5em minus 0.4em\relax British Machine Vision Association, 2018.

\bibitem{raffel2020exploring}
C.~Raffel, N.~Shazeer, A.~Roberts, K.~Lee, S.~Narang, M.~Matena, Y.~Zhou, W.~Li, and P.~J. Liu, ``Exploring the limits of transfer learning with a unified text-to-text transformer,'' \emph{Journal of machine learning research}, vol.~21, no. 140, pp. 1--67, 2020.

\bibitem{google/mt5-base}
\BIBentryALTinterwordspacing
``google/mt5-base,'' 2020. [Online]. Available: \url{https://huggingface.co/google/mt5-base}
\BIBentrySTDinterwordspacing

\bibitem{diederik2014adam}
P.~K. Diederik, ``Adam: A method for stochastic optimization,'' 2014.

\bibitem{lin2004rouge}
C.-Y. Lin, ``Rouge: A package for automatic evaluation of summaries,'' in \emph{Text summarization branches out}, 2004, pp. 74--81.

\bibitem{papineni2002bleu}
K.~Papineni, S.~Roukos, T.~Ward, and W.-J. Zhu, ``Bleu: a method for automatic evaluation of machine translation,'' in \emph{Proceedings of the 40th annual meeting of the Association for Computational Linguistics}, 2002, pp. 311--318.

\end{thebibliography}

\end{document}